\documentclass{article}

\usepackage[final]{neurips_2025}

\usepackage[utf8]{inputenc} 
\usepackage[T1]{fontenc}    
\usepackage{hyperref}       
\usepackage{url}            
\usepackage{booktabs}       
\usepackage{amsfonts}       
\usepackage{nicefrac}       
\usepackage{microtype}      
\usepackage{xcolor}         
\usepackage{amsfonts}       
\usepackage{nicefrac}       
\usepackage{multirow}
\usepackage{wrapfig}
\usepackage{pifont}
\usepackage{amsmath}
\usepackage{graphicx}
\usepackage{floatrow}
\usepackage{caption}
\usepackage{setspace}
\newfloatcommand{figurebox}{figure}[\nocapbeside][\dimexpr(\textwidth-\columnsep)/2\relax]
\newfloatcommand{tablebox}{table}[\nocapbeside][\dimexpr(\textwidth-\columnsep)/2\relax]
\begin{document}
\title{TrackingWorld: World-centric Monocular 3D Tracking of Almost All Pixels}

%

\author{%
Jiahao Lu$^{1}$~\quad Weitao Xiong$^{1,5}$ ~\quad Jiacheng Deng$^{2}$ ~\quad Peng Li$^{1}$ ~\quad 
\\ \textbf{Tianyu Huang}$^{3}$ ~\quad \textbf{Zhiyang Dou}$^{4}$~\quad \textbf{Cheng Lin}$^{6}$ ~\quad \textbf{Sai-Kit Yeung}$^{1}$ ~\quad \textbf{Yuan Liu}$^{1}$\footnotemark[2]\\\\\
$^1$HKUST ~\quad $^2$USTC ~\quad $^3$CUHK ~\quad $^4$HKU ~\quad$^5$XMU ~\quad $^6$MUST\\
\\
\url{https://igl-hkust.github.io/TrackingWorld.github.io/}
}

\renewcommand{\thefootnote}{\fnsymbol{footnote}}
\maketitle
\footnotetext[2]{Corresponding authors}

\maketitle

\begin{abstract}

Monocular 3D tracking aims to capture the long-term motion of pixels in 3D space from a single monocular video and has witnessed rapid progress in recent years. 
However, we argue that the existing monocular 3D tracking methods still fall short in separating the camera motion from foreground dynamic motion and cannot densely track newly emerging dynamic subjects in the videos.
To address these two limitations, we propose TrackingWorld, a novel pipeline for dense 3D tracking of almost all pixels within a world-centric 3D coordinate system. 
First, we introduce a tracking upsampler that efficiently lifts the arbitrary sparse 2D tracks into dense 2D tracks. 
Then, to generalize the current tracking methods to newly emerging objects, we apply the upsampler to all frames and reduce the redundancy of 2D tracks by eliminating the tracks in overlapped regions.
Finally, we present an efficient optimization-based framework to back-project dense 2D tracks into world-centric 3D trajectories by estimating the camera poses and the 3D coordinates of these 2D tracks. 
Extensive evaluations on both synthetic and real-world datasets demonstrate that our system achieves accurate and dense 3D tracking in a world-centric coordinate frame.
\end{abstract}

\section{Introduction}

Estimating long-term motion in dynamic videos remains a persistent challenge in computer vision~\cite{karaev2024cotracker3,wang2023tracking,xiao2024spatialtracker,ngo2024delta}. Fine-grained motion tracking is crucial for understanding object dynamics, modeling camera motion, and facilitating the generation of temporally and geometrically consistent videos~\cite{burgert2025gowiththeflow,gu2025diffusion,qiu2024freetraj}.

In recent years, dense 2D pixel tracking~\cite{harley2022particle,doersch2022tap,doersch2023tapir,karaev2024cotracker3,cho2024local,li2024taptr,zholus2025tapnext,qu2024taptrv3,zhang2025protracker,dong2025online} has emerged as an active research topic, with notable advancements such as CoTrackers~\cite{karaev2024cotracker,karaev2024cotracker3}, which employs transformers to iteratively update 2D tracks and has driven progress in 2D motion analysis. 
This development also motivates many recent works for 3D tracking.
Early 3D tracking works like OmniMotion~\cite{wang2023tracking,song2024track} adopt optimization-based approaches to estimate 3D motion, while subsequent feedforward methods such as SpatialTracker~\cite{xiao2024spatialtracker} and DELTA~\cite{ngo2024delta} leverage extracted features to directly estimate the 3D tracking in a feedforward manner without per-sequence optimization. 
These 3D tracking methods demonstrate substantial potential for downstream applications, including detailed 3D motion analysis and high-fidelity novel view synthesis, highlighting the growing importance of monocular 3D tracking as a critical research frontier.

Upon analyzing all existing 3D tracking methods, we observe that these existing methods still suffer from two noticeable shortcomings. First, these methods~\cite{ngo2024delta,xiao2024spatialtracker,wang2023tracking} cannot distinguish the camera motion and the dynamic object motion. All these methods assume a static camera and just model the 3D flow within the camera coordinate system. However, many downstream tasks like motion analysis or novel-view-synthesis require distinguishing camera motion from the dynamic object motion. Moreover, some recent works~\cite{zhu2024motiongs} also show that explicitly considering camera poses in motion estimation improves the 3D tracking quality. Only some very recent works~\cite{feng2025st4rtrack,zhang2025tapip3d,jin2024stereo4d} try to estimate the 3D tracks in the world-centric coordinate system and enable distinguishing camera motions from dynamic object motions. Estimating camera motion is still challenging for a monocular video containing dynamic objects because only static scenes provide cues for camera pose estimation.

The second shortcoming is that existing methods are mostly limited to tracking sparse pixels in the first frame of the video and cannot track all pixels in all frames (e.g., new objects emerging in the intermediate frames). 
Tracking all pixels brings a huge computational complexity to all tracking methods. Recent works like DELTA~\cite{ngo2024delta} propose to upsample the sparse tracking points with neural networks to produce dense 3D tracks. However, DELTA is still limited to tracking the first frame of the video, and how to estimate the dense 3D tracks for all pixels of all frames still remains an unexplored problem.

\begin{figure}[!t]
    \vspace{-2em}
    \begin{center}
        \includegraphics[width=1\textwidth]{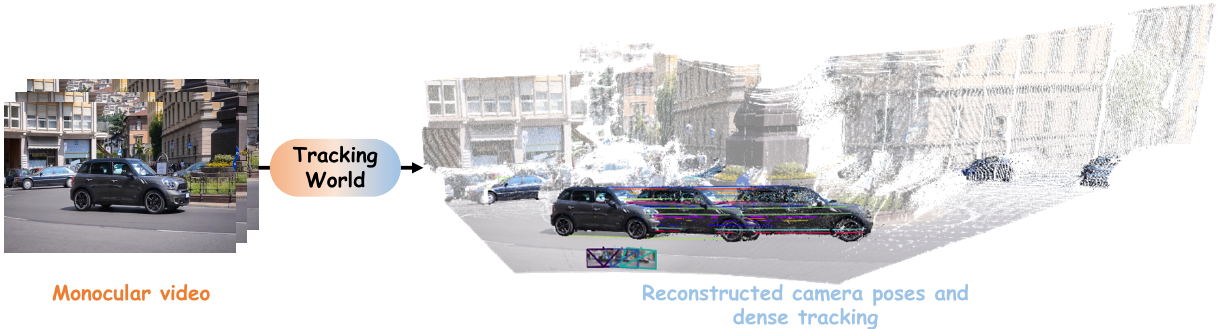}
        \caption{\textbf{TrackingWorld} estimates world-centric dense tracking results from monocular videos. Our model can accurately estimate camera poses and achieve disentangled 3D track modeling of static and dynamic components, not just limited to one foreground dynamic object. We only visualize a subset of foreground dynamic point trajectories and apply a fading color to background static points.
        }
        \label{intro}
    \end{center}
    \vspace{-1.em}
\end{figure}

In this paper, we propose \textbf{TrackingWorld}, a 3D tracking method that enables dense 3D tracking of almost all pixels of all frames from a monocular video within a world-centric coordinate system. ``almost all'' means we filter some noisy and outlier tracks to ensure robustness and accuracy. Specifically, TrackingWorld takes a monocular video and the monocular estimation from foundation models as input, including sparse tracks~\cite{ngo2024delta,karaev2024cotracker3}, depth maps~\cite{piccinelli2024unidepth}, and coarse foreground dynamic masks~\cite{yao2025uni4d}. Then, TrackingWorld produces high quality dense 3D tracks for almost all pixels of the monocular video and the camera poses for every frame. TrackingWorld addresses the above shortcomings with the following strategies.

First, to enable the dense tracking of almost all pixels, we utilize the track upsampler of DELTA~\cite{ngo2024delta} and track every frame iteratively.
We find that the tracking upsampler module of DELTA~\cite{ngo2024delta} is applicable to arbitrary 2D tracks, which are utilized by TrackingWorld to upsample the input sparse 2D tracks to dense 2D tracks. Then, we not only track the pixels of the first frame but also repeat it on all subsequent frames.
To reduce computational complexity, we observe that many regions of subsequent frames have already been seen in the first or previous frames. Therefore, we delete the redundant tracks corresponding to these overlapping regions.

Second, to accurately separate the camera motion from the dynamic object motion, we estimate the 3D tracks and the camera poses from the upsampled dense 2D tracks and the input estimated depth maps.
A key challenge lies in the inaccuracy of the estimated dynamic masks, which often fail to capture dynamic background objects. This limitation leads to suboptimal bundle adjustment interfered by dynamic background objects, ultimately compromising the accuracy of both camera pose estimation and object motion tracking.
Thus, we treat all points in the initial static regions as potentially dynamic but impose an as-static-as-possible constraint for the camera pose estimation, which effectively helps us rule out the dynamic background points for an accurate camera pose estimation. Finally, we utilize the estimated camera poses along with the depth maps to convert all the 2D tracks into 3D tracks in the world coordinate.

To comprehensively evaluate whether our proposed method can effectively achieve dense 3D tracking of almost all pixels across all frames within a world-centric coordinate system, we conduct evaluations from multiple perspectives: 1. Camera pose estimation accuracy; 2. Depth accuracy of the dense 3D tracks; 3. Sparse 3D tracking performance; 4. Accuracy of the dense 2D tracking results.
Our empirical analysis demonstrates that the proposed method yields superior performance across all metrics, confirming its effectiveness in establishing accurate and consistent 3D tracks over time.

\section{Related Work}

\subsection{2D Point Tracking}
The task of tracking arbitrary points~\cite{harley2022particle,doersch2022tap,doersch2023tapir,karaev2024cotracker3,cho2024local,li2024taptr,zholus2025tapnext,qu2024taptrv3,zhang2025protracker,dong2025online} across video frames is first introduced by PIPs~\cite{harley2022particle}, which leverages deep learning to tackle point tracking based on optical flow. Built upon RAFT~\cite{teed2020raft}, PIPs computes inter-frame correlation maps and uses a decoder to iteratively refine tracking results. TAP-Vid~\cite{doersch2022tap} further improves the problem formulation, introducing three standardized benchmarks along with TAP-Net, a dedicated model for point tracking. TAPIR~\cite{doersch2023tapir} advances performance by combining a matching stage with a refinement stage, enhancing tracking accuracy. CoTrackers~\cite{karaev2024cotracker,karaev2024cotracker3} observe that strong correlations exist across different point trajectories, and exploit this insight by training on unrolled sequences over long videos, which significantly improve long-term tracking performance. Drawing inspiration from DETR~\cite{carion2020end}, TAPTR~\cite{li2024taptr} proposes an end-to-end transformer-based architecture, where each point is represented as a query token in the decoder, enabling direct modeling of point dynamics. LocoTrack~\cite{cho2024local} extends traditional 2D correlation features to 4D correlation volumes and introduces a lightweight correlation encoder, achieving better efficiency while preserving accuracy.
\subsection{3D Point Tracking}
While previous works have primarily focused on 2D point tracking, recent research has increasingly focused on 3D point tracking~\cite{wang2023tracking,song2024track,xiao2024spatialtracker,ngo2024delta,wang2024scenetracker,chen2025back,seidenschwarz2024dynomo,cho2025seurat,kasten2024fast,jin2024stereo4d,feng2025st4rtrack,zhang2025tapip3d}. Early 3D tracking methods, such as OmniMotion~\cite{wang2023tracking}, adopt optimization-based approaches to estimate 3D motion. Subsequent work like OmniTrackFast~\cite{song2024track} aims to reduce the optimization time and enhance robustness. More recently, increasing attention has shifted toward feedforward-based methods. For example, SpatialTracker~\cite{xiao2024spatialtracker} represents points in a (u, v, d) coordinate system, combining image-plane coordinates with depth information. It incorporates depth priors and uses a triplane representation to enable effective 3D tracking. Building upon this idea, DELTA~\cite{ngo2024delta} also adopts the UVD coordinate system, but takes a different approach by decoupling appearance and depth correlations. DELTA introduces a coarse-to-fine trajectory estimation strategy, allowing for efficient dense tracking across the entire frame rather than being limited to a sparse set of locations. In contrast to the aforementioned methods that focus on UVD (2.5D) representations, several concurrent works have recently explored 3D tracking in a world-centric coordinate system. St4RTrack~\cite{feng2025st4rtrack} adopts a DUSt3R~\cite{wang2024dust3r}-like framework to establish pairwise correspondences, but this approach may suffer from drift during long-term tracking. TAPIP3D~\cite{zhang2025tapip3d} primarily focuses on sparse tracking and is inherently unable to recover camera motion. In comparison, our method introduces a comprehensive pipeline for dense 3D tracking that can robustly capture newly emerging objects within a world-centric coordinate system.
\subsection{4D Reconstruction}
4D reconstruction~\cite{kopf2021robust,zhang2022structure,li2024megasam,lu2024dn,zhu2024motiongs,zhang2024monst3r,lu2024align3r,cao2025reconstructing,wang2025continuous,yao2025uni4d} aims to recover both camera motion and object motion within a scene. The problem of non-rigid structure
from motion is highly ill-posed. To overcome this limitation, a variety of approaches have been proposed. RobustCVD~\cite{kopf2021robust} refines depth estimation using 3D geometric constraints, while CasualSAM~\cite{zhang2022structure} finetunes a depth network guided by predicted motion masks. 
MegaSaM~\cite{li2024megasam} integrates monocular depth priors and motion probability
maps into a differentiable SLAM paradigm.
Inspired by DUSt3R~\cite{wang2024dust3r}, several data-driven methods such as MonST3R~\cite{zhang2024monst3r}, Align3R~\cite{lu2024align3r}, and Cut3r~\cite{wang2025continuous} adopt 3D point cloud representations to enable full 4D reconstruction. In addition, Uni4D~\cite{yao2025uni4d}, a multi-stage optimization framework, leverages multiple pretrained models to improve reconstruction in dynamic scenes. Its core contribution lies in the use of foundation models to achieve effective separation of static and dynamic elements within the scene. Our method also adopts an optimization-based framework to decouple camera motion and object motion. However, unlike prior works that primarily focus on 4D reconstruction, our approach targets a higher-level task—dense tracking of every pixel—which enables fine-grained correspondence estimation across time. By focusing on dense pixel-level tracking, our method provides a more detailed and temporally consistent understanding of dynamic scenes, making it well-suited for applications such as motion analysis, scene understanding, and video editing~\cite{burgert2025gowiththeflow,gu2025diffusion,qiu2024freetraj}.

\section{Method}
\label{Method}

\begin{figure}[!t]
    \begin{center}
        \includegraphics[width=1\textwidth]{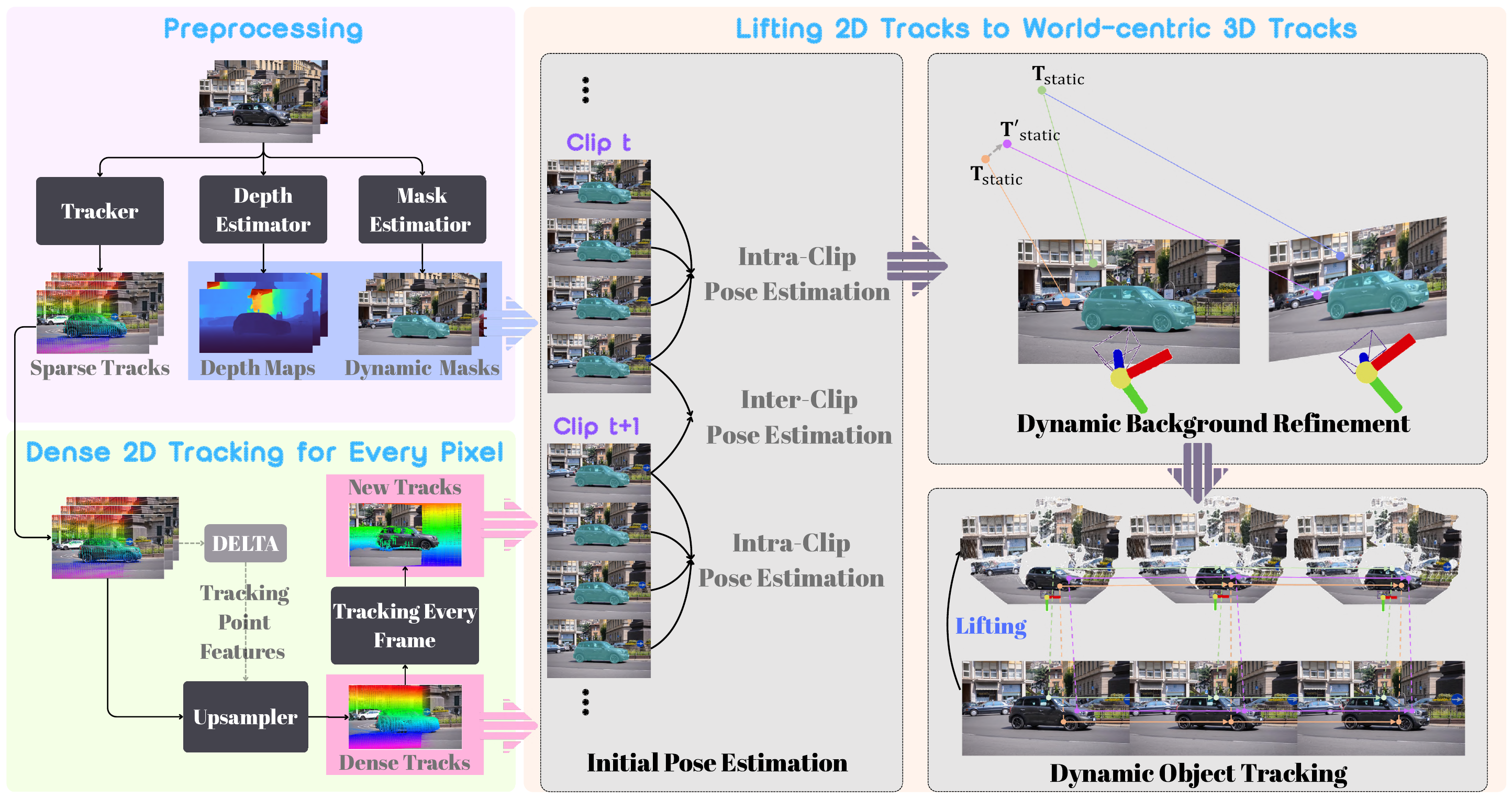}
        \caption{\textbf{Overview}. 
        Given a video sequence, TrackingWorld first generates dense 2D tracking results that are capable of capturing newly emerging objects in the scene. These 2D trajectories are then fed into an optimization-based framework to transform them into a world-centric 3D space. Specifically, we begin by estimating the initial camera poses for each frame at the clip level. We then perform dynamic background refinement to exclude potentially dynamic regions and refine the camera poses. Based on the optimized poses, we finally reconstruct the trajectories of all dynamic regions.
       }
        \label{framework}
    \end{center}
\end{figure}

\subsection{Overview}
Given a video consisting of $T$ frames $\{\mathbf{I}_t \in \mathbb{R}^{\mathrm{H} \times \mathrm{W} \times 3} \mid t=1, \ldots, T\}$, the goal of TrackingWorld is to estimate the corresponding dense 3D trajectories (3D tracks) $\{\mathbf{T}_t \in \mathbb{R}^{\mathrm{M}_t \times 3} \mid t=1, \ldots, T\}$ of almost all pixels, where $\mathrm{M}_t$ denotes the number of tracked points at timestep $t$, along with the camera poses $\{\pi_t\in \mathbb{R}^{3\times 4} | t = 1,...,T\}$. 
Our proposed \textbf{TrackingWorld} framework, illustrated in Fig.~\ref{framework}, achieves this through two main components: first, generating dense 2D tracking results that are capable of following nearly every object in the scene; second, back-projecting these dense 2D tracking results into a world-centric 3D space.

\paragraph{Preprocessing with vision foundation models.}
For the monocular video, we first preprocess it with a 2D tracking model, a foreground dynamic mask estimation module, and a monocular depth estimation module to get a set of 2D tracks, dynamic masks, and depth maps for all frames. For the 2D tracking model, we choose the CoTrackerV3~\cite{karaev2024cotracker3} or the 2D tracking part of DELTA~\cite{ngo2024delta}. For the dynamic mask estimation method, we follow Uni4D~\cite{yao2025uni4d} to apply VLM~\cite{achiam2023gpt} and Grounding-SAM~\cite{sam,liu2024grounding} to segment out foreground dynamic objects. Alternatively, we could also choose the SegmentAnyMotion~\cite{huang2025segment} to get dynamic masks. For the depth estimation, we choose UniDepth~\cite{piccinelli2024unidepth}. Note that all these predictions are not required to be accurate, and we may also adopt other foundation models for this purpose. 

\subsection{Dense 2D Tracking for Every Pixel}
\label{dtep}
In this section, our target is to achieve dense 2D tracking of almost any pixels in the video. We achieve this through two modules: First, we lift the input sparse 2D tracks for a frame to dense 2D tracks; Second, we repeat tracking on every frame and eliminate the overlapped redundant 2D tracks.

\paragraph{Sparse to dense tracks.}
Given the sparse 2D tracks $\mathbf{P}_{\text{sparse}}\in \mathbb{R}^{(\frac{H}{s}\times \frac{W}{s})\times T \times 2}$ for a specific frame, this module aims to lift the sparse 2D tracks to dense 2D tracks $\mathbf{P}_{\text{dense}}\in \mathbb{R}^{(H \times W) \times T \times 2}$. $s$ means the downsampled factor. We achieve this by utilizing the upsampler module of DELTA~\cite{ngo2024delta}. The upsampler module takes the sparse tracks $\mathbf{P}_{\text{sparse}}$ and features defined on the sparse 2D tracks $\mathbf{F}_{\text{sparse}}\in  \mathbb{R}^{\frac{H}{s}\times \frac{W}{s}\times T \times C}$, where $C$ is the feature dimension, as inputs and predicts a weight matrix $\mathbf{W}\in \mathbb{R}^{(\frac{H}{s}\times \frac{W}{s})\times (H\times W)}$. Then, the upsampled dense 2D tracks are
\begin{equation}
    \mathbf{P}_{\text{dense}} = \mathbf{W}^{T} \mathbf{P}_{\text{sparse}},
\end{equation}
where $\mathbf{W}$ actually only correlates a dense track in $\mathbf{P}_{\text{dense}}$ with its neighboring 2D tracks in $\mathbf{P}_{\text{sparse}}$. We find that this upsampler module is not only compatible with DELTA's 2D tracks but also generalizes to arbitrary 2D tracks, so we adopt it here to upsample the arbitrary input sparse 2D tracks into dense 2D tracks for a specific frame.

\paragraph{Tracking every frame.} 
Based on the above upsampler, we further enable tracking of almost all pixels of all frames. To achieve this, we conduct 2D tracking and the sparse-to-dense upsampling on all frames in the video. However, this leads to a large redundancy on the tracking points because most regions are already seen in the previous frames, while only a few regions are new. To avoid wasting computation on these redundant 2D tracks in the subsequent computation, if a pixel resides near the tracking trajectory of arbitrary visible previous 2D tracks, then we discard the pixel. More details can be found in A.6 of the supplementary material.

\subsection{Lifting 2D Tracks to World-centric 3D Tracks}
\label{bp}
In this module, we will estimate the camera poses of all frames and lift the dense 2D tracks estimated by the previous section to 3D tracks in the world-centric coordinate system. We achieve this through the following three steps: First, we utilize the input estimated coarse dynamic masks and estimate the camera poses using only the coarse static regions. However, the dynamic masks are usually not accurate enough, and some dynamic objects in the background still remain. Second, we utilize an as-static-as-possible constraint to further improve the camera pose estimation and find out the dynamic objects in the background. Finally, we transform all 2D tracks within the dynamic regions into 3D tracks in the world-centric coordinate system.

\paragraph{Initial camera pose estimation.}  
In this step, we want to estimate per-frame camera poses \( \{\pi_t \in \mathrm{SE}(3)\} \) from the 2D tracks on static regions and the estimated depth maps. 
We first utilize the input dynamic foreground masks to select 2D tracks $\mathbf{P}_{\text{static}}\in \mathbb{R}^{N_{\text{static}}\times T \times 2}$ on these static regions. Then, for each static 2D track, we unproject its location at timestep \( t_1 \) into the 3D space using the monocular depth map $\mathbf{D}_{\text{static}} \in \mathbb{R}^{N_{\text{static}} \times T \times 1}$. The resulting 3D points are subsequently reprojected into the image plane at timestep \( t_2 \) using the camera poses. Then, we define the projection loss to optimize the camera poses
\begin{equation}
\mathcal{L}_{\text{proj}} = \sum_i^{N_{\text{inliers}}} \sum_{t_1}^{T}\sum_{t_2}^{T} \|\pi_{t_2} \pi_{t_1}^{-1}(\mathbf{P}_{\text{static}}(i,t_1), \mathbf{D}_{\text{static}}(i,t_1))-\mathbf{P}_{\text{static}}(i,t_2)\|_2^2,
\end{equation}
where $\pi_t(\cdot)$ means project with the camera pose on timestep $t$, 
and $\mathbf{P}_{\text{static}}(i,t)\in\mathbb{R}^2$ means the position of the $i$-th static 2D track on timestep $t$, $\mathbf{D}_{\text{static}}(i,t)$ means the depth value for the $i$-th static point on time step $t$, and
$N_{\text{inliers}}$ denotes the number of static 2D tracks whose projection errors fall within the threshold $\tau$.

To further improve the computational efficiency for camera pose estimation, we first divide the entire video into \( C \) clips and estimate camera poses within each clip in parallel. 
After estimating camera poses within each clip, we estimate the pose between clips to merge the camera poses together.

\paragraph{Dynamic background refinement.} 
The foreground dynamic object masks are usually not accurate enough, so some dynamic objects in the background still exist in the assumed ``static'' regions and prevent us from accurately estimating camera poses.
Thus, we further refine the camera pose estimation by treating these static regions as dynamic and introducing an as-static-as-possible constraint. 

Specifically, each static 2D track corresponds to a unique 3D point in the world-centric coordinate system, denoted as $\mathbf{T}_{\text{static}} \in \mathbb{R}^{N_{\text{static}} \times 3}$. We initialize $\mathbf{T}_{\text{static}}$ by back-projecting the static 2D tracks using the depth estimated by UniDepth and the camera poses obtained from the previous stage: Initial camera pose estimation. Notably, for each 2D track $\in \mathbb{R}^{T \times 2}$, we only back-project the visible timesteps and take the average of the resulting 3D points.
To better model the potentially dynamic regions that are not accurately segmented, we introduce an additional object motion term \( \mathbf{O}_{\text{static}} \in \mathbb{R}^{\mathrm{N}_{\text{static}} \times T \times 3} \), which captures residual object motions over time. With this term, the time-dependent world-centric static tracking becomes
\begin{equation}
\mathbf{T}_{\text{static}}'(i,t) = \mathbf{T}_{\text{static}}(i) + \mathbf{O}_{\text{static}}(i,t),
\end{equation}
where $\mathbf{T}_{\text{static}}'(i,t)\in \mathbb{R}^3$ means the 3D coordinate of the $i$-th static point at timestep $t$ and $\mathbf{O}_{\text{static}}(i,t)$ is the corresponding 3 dimensional offset. 
We then jointly optimize the camera poses \( \pi_t \) and the static 3D coordinates \( \mathbf{T'}_{\text{static}} \) using a bundle adjustment loss:
\begin{equation}
    \mathcal{L}_{\text{ba}} = \sum_{i=1}^{N_{\text{static}}} \sum_{t=1}^{T} \left\| \pi_t(\mathbf{T}_{\text{static}}'(i,t)) - \mathbf{P}_{\text{static}}(i,t) \right\|_2^2,
    \label{eq:proj}
\end{equation}
where \( \mathbf{P}_{\text{static}}(i,t) \) is the observed 2D projection of the \( i \)-th track at timestep \( t \).
In addition to the bundle adjustment loss, we also compute a depth consistency loss $\mathcal{L}_{\text{dc}}$ to enforce the consistency between the projected depth maps from $T'_{\text{static}}$ and the estimated monocular depth maps, as introduced in the supplementary material. 
To ensure that residual motion remains minimal for genuinely static regions, we regularize the offset $\mathbf{O}_{\text{static}}$ with an as-static-as-possible constraint
\begin{equation}
    \mathcal{L}_{\text{asap}}=\sum_{i,t}\|\mathbf{O}_{\text{static}}(i,t)\|_1,
\end{equation}
where we minimize the L1 norms of offsets to make all points as static as possible. This $\mathcal{L}_{\text{asap}}$ enables the accurate camera estimation and also models the dynamics of background objects.

\paragraph{Dynamic object tracking.} 
In this step, our target is to lift the 2D tracks of dynamic regions to 3D tracks. We also include the dynamic background points with $\|\mathbf{O}_{\text{static}}(i,\cdot)\|_2\ge \varepsilon$ here as the dynamic 3D tracks.
For these dynamic 3D tracks, we directly represent their 3D coordinates by $\mathbf{T}_{\text{dynamic}} \in \mathbb{R}^{N_{\text{dynamic}} \times T \times 3}$. Similar to the 3D static tracks, we initialize the dynamic 3D tracks by back-projecting them using the depths predicted by UniDepth and the camera poses refined in the second stage. Based on $\mathbf{T}_{\text{dynamic}}$, we also compute the projection loss in Eq.~\ref{eq:proj}, the depth consistency loss $\mathcal{L}_{\text{dc}}$, as-rigid-as-possible loss $\mathcal{L}_{\text{arap}}$~\cite{wang2024shape,yao2025uni4d}, and a temporal smoothness loss $\mathcal{L}_{\text{ts}}$~\cite{yao2025uni4d}. All the details of these loss terms are included in the supplementary material. The final outputs are the dynamic 3D tracks $\mathbf{T}_{\text{dynamic}}$, static 3D tracks $\mathbf{T}_{\text{static}}'$, and the camera poses $\pi_t$.

\textbf{Discussion}. The tracking module TrackingWorld differs from previous 3D tracking methods, DELTA~\cite{ngo2024delta} and SpatialTracker~\cite{xiao2024spatialtracker} by explicitly estimating the camera poses, which enables the estimation of 3D tracks in the world-centric coordinate system. The explicit separation between camera motion and object motion also improves the quality of 3D tracking because of the better decomposition, as demonstrated by experimental results in Tab.~\ref{table: Sparse}. In comparison with the existing dynamic video camera pose estimation methods, like Uni4D~\cite{yao2025uni4d}, we do not just assume a single dynamic foreground object but also model the background object motion in the camera pose estimation for a better performance. Instead of simply discarding these dynamic background objects, we also track their 3D points in the world-centric coordinate system, enabling tracking almost all pixels.

\section{Experiment}
\label{Experiment}
\subsection{Implementation details}
\label{Implementation}
All experiments are conducted on an RTX 4090 GPU. We use CoTrackerV3~\cite{karaev2024cotracker3} and DELTA~\cite{ngo2024delta} to obtain dense tracking results, and adopt UniDepth~\cite{yao2025uni4d} as the depth prior. The entire framework takes $\sim$20 minutes to produce dense world-centric 3D tracking results for a 30-frame video. All baseline methods are run on the datasets using their official implementations and default hyperparameters. More details about hyperparameters can be found in the supplementary materials.

\subsection{Quantitative comparisons}
To demonstrate the capability of our method in dense 3D tracking within a world-centric coordinate system, we evaluate the following performance: 1. Camera pose estimation accuracy; 2. Depth accuracy of dense 3D tracks; 3. Sparse 3D tracking performance; 4. Dense 2D tracking performance.

\subsubsection{Camera pose estimation results}
\paragraph{Benchmarks and metrics.} We evaluate camera pose estimation performance on three dynamic datasets: Sintel~\cite{butler2012naturalistic}, Bonn~\cite{palazzolo2019refusion}, and TUM-D~\cite{sturm2012benchmark}. For all three datasets, we adopt the same settings as MonST3R~\cite{zhang2024monst3r}. Following~\cite{chen2024leap,zhao2022particlesfm,teed2024deep}, we report three ATE $\downarrow$ (Absolute Trajectory Error), RTE $\downarrow$ (Relative Translation Error), and RRE $\downarrow$ (Relative Rotation Error). 
ATE measures the deviation between estimated and ground truth trajectories after alignment. RTE and RRE evaluate the average local translation and rotation errors over consecutive pose pairs, respectively.

\paragraph{Comparison with existing methods.} Tab.~\ref{pose_estimation} presents the quantitative comparison between our method and existing approaches. To recover the camera pose, we first obtain dense tracking results, followed by optimization process that refines the camera pose and world-centric dense tracking. As shown in the table, regardless of whether the dense tracking is derived from DELTA~\cite{ngo2024delta} or CoTrackerV3~\cite{karaev2024cotracker3}, our method consistently achieves more accurate pose estimation than previous approaches across all three datasets.

\begin{table*}[!ht]
  \begin{center}
    \footnotesize
    \setlength\tabcolsep{2.5pt}
    \caption{\textbf{Camera pose estimation results.} We evaluate our model on three datasets: Sintel, Bonn, and TUM-D. \textbf{Best} results are highlighted. $\ddagger$ means using ground truth camera intrinsics as input. * means reproduced by 2D tracks from DELTA, the same as ``Ours(DELTA)''.
}
    \label{pose_estimation}
    \resizebox{\textwidth}{!}{
    \begin{tabular}{ll|ccc|ccc|ccc}
      \toprule
       \multirow{2}{*}{Category} &\multirow{2}{*}{Method} &   \multicolumn{3}{c|}{Sintel} &\multicolumn{3}{c|}{Bonn}&\multicolumn{3}{c}{TUM-D} \\
       & & ATE $\downarrow$&RTE$\downarrow$   & RRE$\downarrow$& ATE $\downarrow$& RTE$\downarrow$   & RRE$\downarrow$& ATE $\downarrow$&RTE$\downarrow$   & RRE$\downarrow$ \\
       \midrule
       
     &  DROID-SLAM$\ddagger$~\cite{teed2021droid}&0.175 &0.084& 1.912 &/&/&/ &/&/&/\\
    Pose only&   DPVO$\ddagger$~\cite{teed2024deep}&0.115 &0.072 &1.975 &/&/&/ &/&/&/\\
    & COLMAP~\cite{schoenberger2016sfm} &0.559&0.325&7.302&/&/&/& 0.076&0.059&7.689  \\
    \midrule
     &Robust-CVD~\cite{kopf2021robust}&0.360 &0.154 &3.443&/&/&/& 0.153& 0.026 &3.528 \\
    &DUSt3R~\cite{wang2024dust3r}&0.601&	0.214	&11.43&0.046&	0.014&	1.836&0.083& 0.017 &3.567\\
    Joint depth&MonST3R~\cite{zhang2024monst3r}&0.111&0.044&0.780&0.029	&0.007	&0.612&0.063&	0.009&	1.217\\
    \& pose&Align3R~\cite{lu2024align3r} (Depth Pro~\cite{bochkovskii2024depthpro})&0.128&0.042&0.432&0.023&0.007&0.620&0.027	&0.018&0.446\\
    &Uni4D*~\cite{yao2025uni4d}&0.116&0.046&0.603&0.017&0.006&0.561&0.039&0.007&0.434\\
     & Ours (CoTrackerV3~\cite{karaev2024cotracker3}) &0.103&0.039&0.439&0.016&\textbf{0.005}& \textbf{0.561}&\textbf{0.014}&\textbf{0.005}&0.338\\
    & Ours (DELTA~\cite{ngo2024delta}) &\textbf{0.088}&\textbf{0.035}&\textbf{0.410}&\textbf{0.016}&0.005& 0.564&0.016&0.005&\textbf{0.333}\\

      \bottomrule
    \end{tabular}
    }
     \vspace{-1.2em}
  \end{center}
\end{table*}

\subsubsection{Depth accuracy of the dense 3D tracks}
\paragraph{Benchmarks and metrics.} Since our method does not aim to optimize 2D tracking accuracy directly, but rather focuses on how to transform 2D tracking into dense world-centric tracking, we evaluates the accuracy of the camera-centric depth for each tracked point. Specifically, we compare the predicted depth with the ground-truth depth only for tracked points that lie within the image bounds. As multiple tracked points may track the same pixel, we retain the one with the smaller depth value for evaluation, assuming it more likely corresponds to the visible surface. Similar to the camera pose benchmark, we evaluate on the same datasets and under identical settings: Sintel, Bonn, and TUM-D. Following prior works~\cite{hu2024depthcrafter,zhang2024monst3r}, we align the estimated dense tracking depth with the ground truth using a single scale and shift before computing the evaluation metrics. We primarily report two metrics: Abs Rel $\downarrow$ (absolute relative error) and the percentage of inlier points with $\delta < 1.25 \uparrow$.
\paragraph{Comparison with existing methods.} Tab.~\ref{table:dense_trackingdepth} reports the results of dense tracking depth estimation. Thanks to our optimization-based bundle adjustment, which enforces strong 3D geometric consistency, the estimated tracking depth is significantly improved across all datasets.

\begin{table}[!ht]
  \begin{center}
    \footnotesize
    \setlength\tabcolsep{1.5pt}
    \caption{\textbf{Depth accuracy of the dense 3D tracks.} \textbf{Best} results are highlighted.}
    \label{table:dense_trackingdepth}
    \scalebox{1}{\begin{tabular}{l|c|cc|cc|cc}
      \toprule
       \multirow{2}{*}{Method}&\multirow{2}{*}{Depth Prior}&\multicolumn{2}{c|}{Sintel} &\multicolumn{2}{c|}{Bonn} &  \multicolumn{2}{c}{TUM-D}    \\
       && Abs Rel $\downarrow$& $\delta<1.25 \uparrow$   & Abs Rel $\downarrow$& $\delta<1.25 \uparrow$& Abs Rel $\downarrow$& $\delta<1.25 \uparrow$\\
       \midrule

        DELTA~\cite{ngo2024delta}&ZoeDepth~\cite{bhat2023zoedepth}&0.814&46.1&0.168&88.5 &0.239&70.5\\
        DELTA~\cite{ngo2024delta}&Depth Pro~\cite{bochkovskii2024depthpro}&0.813&50.7&0.160&90.6 &0.222&78.4\\
        DELTA~\cite{ngo2024delta}&Unidepth~\cite{piccinelli2024unidepth}&0.636&63.1&0.153&90.5 &0.178&85.6\\
        Ours (CoTrackerV3~\cite{karaev2024cotracker3})&Unidepth~\cite{piccinelli2024unidepth}&0.219&73.1&\textbf{0.054}&97.2&0.089&91.5\\
        Ours (DELTA~\cite{ngo2024delta})&Unidepth~\cite{piccinelli2024unidepth}&\textbf{0.218}&\textbf{73.3}&0.058&\textbf{97.3}&\textbf{0.084}&\textbf{92.3}\\
        
      \bottomrule
    \end{tabular}}
     \vspace{-1.2em}
  \end{center}
\end{table}

\subsubsection{Sparse 3D tracking results}
\paragraph{Benchmarks and Metrics.}
To evaluate the performance of 3D sparse tracks, we conduct experiments on two datasets, ADT~\cite{pan2023aria} with moving cameras, and PStudio~\cite{joo2015panoptic} with static cameras. For each dataset, the video subsets for evaluation are selected at fixed intervals: for ADT, we sample one video every 100 videos, and for PStudio, one video every 20 videos. The sparse 3D tracking result are evaluated in camera coordinates. As for evaluation metrics, we adopt Average Jaccard (AJ), which jointly evaluates the accuracy of both spatial position and occlusion estimation, serving as a comprehensive indicator of tracking quality; APD$_{3D}$ ($< \delta_{\text{avg}}$) which measures the average percentage of tracked points whose errors fall within a given threshold $\delta$, reflecting geometric accuracy; Occlusion Accuracy (OA) which evaluates the precision of occlusion state prediction across frames. 

\paragraph{Comparison with existing methods.} 
Since our method primarily focuses on dense tracking, we maintain the optimization of dense tracking results even when evaluating sparse tracking performance. To this end, we sample evaluation points from the optimized dense tracking set. As shown in Tab.~\ref{table: Sparse}, our method achieves higher 3D geometric consistency in tracking. For scenes with camera motion (ADT), the explicit separation between camera motion and object motion leads to significant improvements in both AJ and APD$_{3D}$. In contrast, for scenes with static cameras (PStudio), the benefits from geometric optimization are relatively limited, resulting in smaller performance gains. It is worth noting that OA mainly evaluates the visibility accuracy of tracking points. Since we directly adopt the visibility maps predicted by DELTA/CoTrackerV3, the OA scores remain consistent with those of DELTA/CoTrackerV3.

\begin{figure}[!tp]
\vspace{-1.2em}
  \begin{floatrow}[2]
    \tablebox{\caption{\textbf{Sparse 3D tracking results.} ``Feed.'' means feedforward methods while ``Optim'' means optimization-b ased method.}\vspace{-1.2em}}{%
    \label{table: Sparse}
    \scalebox{0.65}{
    \setlength\tabcolsep{1.pt}
    \begin{tabular}{ll|ccc|ccc}
      \toprule
       \multirow{2}{*}{Category}&\multirow{2}{*}{Method}&\multicolumn{3}{c|}{ADT} &  \multicolumn{3}{c}{PStudio}  \\
       && AJ$\uparrow$& APD$_{3D}$ $\uparrow$   & OA $\uparrow$& AJ$\uparrow$& APD$_{3D}$ $\uparrow$   & OA $\uparrow$\\
       \midrule
        &CoTrackerV3~\cite{karaev2024cotracker3}+Uni~\cite{piccinelli2024unidepth}&13.6&21.3&88.5&14.1&22.8&\textbf{87.7}\\
        
        Feed.&SpatialTracker~\cite{xiao2024spatialtracker}&14.3&22.3&\textbf{91.5}&13.8&23.7&79.5\\
        &DELTA~\cite{ngo2024delta}&15.3&22.9&90.1&15.1&24.6&75.7\\
        \midrule
        &OmniTrackFast~\cite{song2024track}&8.6&18.2&63.9&6.4&12.2&81.8\\
        Optim.&Ours (CoTrackerV3~\cite{karaev2024cotracker3})&22.5&31.5&88.5&14.2&24.0&\textbf{87.7}\\
        &Ours (DELTA~\cite{ngo2024delta})&\textbf{23.4}&\textbf{32.2}&90.1&\textbf{15.1}&\textbf{25.6}&75.7\\

      \bottomrule
    \end{tabular}

    }}
  \tablebox{\caption{\textbf{Long-range optical flow results.}}\vspace{-1.2em}}{%
  \label{table:optical}
    \scalebox{0.7}{\begin{tabular}{l|cc|cc}
      \toprule
       \multirow{2}{*}{Method}&\multicolumn{2}{c|}{CVO-Clean} &  \multicolumn{2}{c}{CVO-Final}  \\
       & EPE$\downarrow$  & IoU$\uparrow$& EPE$\downarrow$ & IoU$\uparrow$\\
       \midrule
        RAFT~\cite{teed2020raft}& 2.48& 57.6& 2.63& 56.7\\
        CoTracker~\cite{karaev2024cotracker} &1.51 & 75.5& 1.52 &75.3\\
        SpatialTracker~\cite{xiao2024spatialtracker}& 1.84 &68.5 &1.88& 68.1\\
        DOT-3D~\cite{le2024dense}& 1.33  &79.0  &1.38 &78.8\\
        DELTA~\cite{ngo2024delta}&\textbf{1.14}&78.9& 1.39& 78.2\\
        CoTrackerV3~\cite{karaev2024cotracker3} + Up &1.24&\textbf{80.9}&\textbf{1.35}&\textbf{80.6}\\
        
      \bottomrule
    \end{tabular}}
  }
  \end{floatrow}
    \vspace{-1.2em}
\end{figure}

\subsubsection{Accuracy of dense 2D tracks}

\paragraph{Benchmarks and Metrics.}
We evaluate the dense 2D tracking performance on the CVO~\cite{wu2023accflow} test set, which consists of two subsets: CVO-Clean and CVO-Final, with the latter incorporating motion blur. Each subset contains approximately 500 videos with 7 frames. For evaluation, we adopt the following metrics: the end-point error (EPE) between the predicted and ground-truth optical flows for all points, and the intersection-over-union (IoU) between the predicted and ground-truth occluded regions in visible masks.
\paragraph{Comparison with existing methods.}
To verify the accuracy of the 2D dense tracks generated by the upsampler module (Up) introduced in Sec.~\ref{dtep}, we conduct additional long-range optical flow experiments, as shown in Tab.~\ref{table:optical}. The results demonstrate that the upsampler module generalizes well to other 2D trackers, such as CoTrackerV3, achieving comparable performance with DELTA.

\subsection{Qualitative results}
\begin{figure}[!t]
    \vspace{-2.em}
    \begin{center}
        \includegraphics[width=1\textwidth]{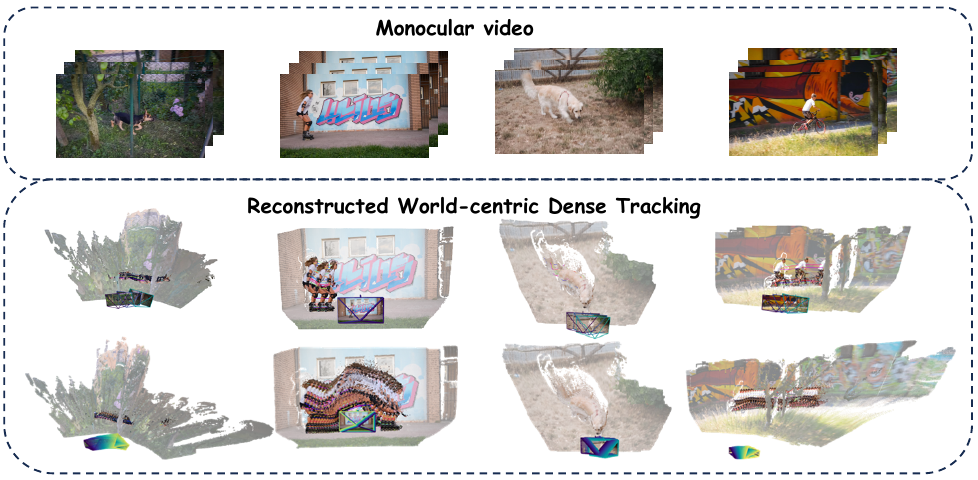}
        \caption{\textbf{Qualitative results on DAVIS dataset.} Our method can output both reliable camera trajectories and world centric dense tracking. The second row visualizes 3D tracking results on temporally spaced keyframes, while the third row shows complete tracks across continuous frames.
        }
        \label{vis}
    \end{center}
    \vspace{-1.2em}
\end{figure}

Fig.~\ref{vis} qualitatively visualizes the world-centric dense tracking results produced by our method on the DAVIS~\cite{perazzi2016benchmark} dataset. For each video sequence, the second row displays 3D tracking results on temporally spaced keyframes, making the changes in object trajectories more perceptible while avoiding visual clutter. The third row presents continuous 3D tracks across all frames, offering a comprehensive view of motion consistency and trajectory completeness. As discussed in Sec.~\ref{bp}, by separating dynamic and static elements, we can generate stable tracking results for both the static background and dynamic objects.

\subsection{Ablation study}

\begin{figure}[!htp]
\vspace{-0.5em}
  \begin{floatrow}[2]
    \tablebox{\caption{\textbf{Ablation study on Sintel dataset.}}}{%
    \label{table: ablation}
    \scalebox{0.8}{
     \setlength\tabcolsep{1.pt}
    \scalebox{1}{\begin{tabular}{c|ccc|cc}
      \toprule
        &\multicolumn{5}{c}{Sintel} \\
        Setting& ATE $\downarrow$&RTE$\downarrow$   & RRE$\downarrow$& Abs Rel $\downarrow$& $\delta<1.25 \uparrow$ \\
       \midrule
        w/o T.E.F &0.171&0.047&0.748&/&/\\
         w/o pose-init.&0.659&0.153&1.382&0.230&72.4 \\
         w/o D.O.T&0.088&0.035&0.410&0.468&73.0\\
         \midrule
        w/o $N_{\text{inliers}}$ &0.089&0.035&0.414&0.220&72.9\\
        w/o $\mathbf{O}_{\text{static}}$&0.092&0.036&0.459&0.224&72.6\\
        w/o $\mathcal{L}_{\text{dc}}$&0.093&0.036&0.441&0.234&71.2\\
        \midrule
        Full&\textbf{0.088}&\textbf{0.035}&\textbf{0.410}&\textbf{0.218}&\textbf{73.3}\\
      \bottomrule
    \end{tabular}}

    }}
  \figurebox{\caption{\textbf{Effectiveness of $\mathbf{O}_{\text{static}}$.} Key regions are highlighted in red.}}{%
  \includegraphics[width=0.48\textwidth]{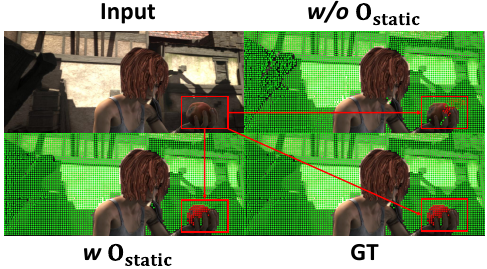}
  \label{figure:omr}
 
  }
  \end{floatrow}
  \vspace{-1.5em}
\end{figure}
\paragraph{Ablation study on the different components.} As shown in Tab.~\ref{table: ablation}, we conduct ablation studies to validate our major design choices. Specifically, the different configurations are as follows: 1) without tracking every frame (w/o T.E.F): In this setting, we only track from the first frame, which leads to the loss of many critical cues for pose estimation, thereby resulting in a significant performance drop. 2) without initial camera pose estimation (w/o pose-init.): We observe that under this setting, it becomes difficult to jointly optimize both camera poses and 3D tracks effectively — a good initialization of the camera poses is necessary to achieve satisfactory results. 3) without dynamic object tracking (w/o D.O.T): In this setting, the depths of dynamic tracks are directly obtained from UniDepth predictions without further refinement. As shown in the table, optimizing dynamic tracks is crucial for achieving better performance. 4) without selecting the inliers whose reprojection errors are within a threshold $\tau$ (w/o $N_{\text{inliers}}$): By filtering all static points and optimizing with nearly static points, we can effectively reduce the influence of outlier trajectories and obtain more accurate camera poses. 5) without the object motion term $\mathbf{O}_{\text{static}}$ (w/o $\mathbf{O}_{\text{static}}$): 
We do not consider the dynamic objects in the assumed ``static'' background and directly optimize the camera poses with all ``static'' background.
We show the projected background ``static'' points (in green and red dots) in Fig.~\ref{figure:omr}. As we can see, the ``apple'' (in the red dots) is considered a background static region but is actually dynamic. Without modeling dynamic points in the background, the points of the apple are incorrectly projected onto incorrect regions.
6) without the depth consistency loss (w/o $\mathcal{L}_{\text{dc}}$): $\mathcal{L}_{\text{dc}}$ can enforce the consistency between the projected depths and the estimated monocular depths, which helps suppress abnormal depth estimations to some extent.

\paragraph{Ablation on different depth estimation models.}
We conducted an ablation study using three commonly used monocular depth estimation models: ZoeDepth~\cite{bhat2023zoedepth}, Depth Pro~\cite{bochkovskii2024depthpro}, and UniDepth~\cite{piccinelli2024unidepth}. 
For all experiments, we fixed the tracking component to DELTA~\cite{ngo2024delta} and evaluated both the camera pose estimation accuracy and the depth accuracy of the dense 3D tracks on the Sintel dataset. As shown in Tab.~\ref{table: different_depth}, our method consistently improves over raw depth predictions across all depth models, especially in downstream tasks such as camera pose estimation. This demonstrates that our pipeline is robust to different depth estimation backbones.

\begin{table}[!ht]
\centering
\footnotesize
\caption{Ablation study on different depth estimation models.}
\vspace{0.5em}
\label{table: different_depth}
\begin{tabular}{lccccc}
\toprule
Method & ATE $\downarrow$ & RTE $\downarrow$ & RPE $\downarrow$ & Abs Rel $\downarrow$ & $\delta<1.25 \uparrow$ \\
\midrule
ZoeDepth & / & / & / & 0.814 & 46.1 \\
Depth Pro & / & / & / & 0.813 & 50.7 \\
UniDepth & / & / & / & 0.636 & 63.1 \\
\midrule
Ours (ZoeDepth) & 0.093 & 0.038 & 0.418 & 0.236 & 72.1 \\
Ours (Depth Pro) & 0.101 & 0.036 & 0.434 & 0.228 & 72.6 \\
Ours (UniDepth) & \textbf{0.088} & \textbf{0.035} & \textbf{0.410} & \textbf{0.218} & \textbf{73.3} \\
\bottomrule
\end{tabular}
\end{table}

\paragraph{Ablation on dynamic mask segmentators.}
As shown in Tab.~\ref{table: dynamic}, we also evaluate different sources of dynamic mask segmentations and observe comparable performance, further demonstrating the robustness of our pipeline.
\begin{table}[!ht]
\centering
\footnotesize
\caption{Ablation study on different dynamic mask segmentators.}
\vspace{0.5em}
\label{table: dynamic}
\begin{tabular}{lccccc}
\toprule
Method & ATE $\downarrow$ & RTE $\downarrow$ & RPE $\downarrow$ & Abs Rel $\downarrow$ & $\delta<1.25 \uparrow$ \\
\midrule
Ours + VLM + GroundingSAM & \textbf{0.088} & \textbf{0.035} & 0.410 & \textbf{0.218} & 73.3 \\
Ours + Segment Any Motion & 0.093 & 0.041 & \textbf{0.379} & 0.224 & \textbf{73.3} \\
\bottomrule
\end{tabular}
\end{table}

\paragraph{Necessity of the 2D upsampler module.}
The 2D upsampler is crucial for achieving efficient dense tracking. Directly predicting dense 2D correspondences (e.g., using CoTrackerV3~\cite{karaev2024cotracker3}) is computationally expensive and memory-intensive, with no clear accuracy gain. To validate this, we compare CoTrackerV3 with and without our upsampler on the CVO-Clean dataset (7-frame sequences). As shown in Tab.~\ref{table: upsampler}, the upsampler improves both accuracy (lower EPE, higher IoU) and drastically reduces runtime (approximately $12\times$ speed-up). This supports our design choice. 

\begin{table}[!ht]
\centering
\footnotesize
\caption{Ablation on the 2D upsampler module.}
\label{table: upsampler}
\vspace{0.5em}
\begin{tabular}{lccc}
\toprule
Method & EPE $\downarrow$ & IoU $\uparrow$ & Avg. Time (min) $\downarrow$ \\
\midrule
CoTrackerV3 & 1.45 & 76.8 & 3.00 \\
CoTrackerV3 + Up & \textbf{1.24} & \textbf{80.9} & \textbf{0.25} \\
\bottomrule
\end{tabular}
\end{table}

\section{Conclusion}

In this paper, we propose TrackingWorld, a novel method for dense 3D tracking of almost all pixels of all frames from a monocular video within a world-centric coordinate system. The key idea of TrackingWorld is to explicitly disentangle camera motion from foreground dynamic motion while densely tracking newly emerging objects. We first introduce a tracking upsampler to densify sparse 2D tracks and apply it to capture newly emerging objects. Finally, we design an efficient optimization-based framework to lift dense 2D tracks into consistent 3D world-centric trajectories. Extensive evaluations across multiple dimensions demonstrate the effectiveness of our system.

{
\small
\bibliographystyle{unsrt}
\bibliography{egbib}
}



\newpage
\appendix
\section{Appendix / supplemental material}
\subsection{More implementation details}
For hyperparameters, the stride $s$ in Sec.~3.2 follows the same setting as in DELTA and is set to 4. The temperature $\tau$ is set to 0.1. Each video clip contains 5 frames. The perturbation parameter $\varepsilon$ is also set to 0.1.

\subsection{Extended explanation of optimization losses}
In dynamic background refinement (Stage 2), the depth consistency loss $\mathcal{L}_{\text{dc}}$ is defined as
\begin{equation}
    \mathcal{L}_{\text{dc}} = \sum_{i=1}^{N_{\text{static}}} \sum_{t=1}^{T} \left\| d(\mathbf{T}_{\text{static}}'(i,t), \pi_t) - \mathbf{D}_{\text{static}}(i,t) \right\|_2^2,
    \label{eq:dc}
\end{equation}
where \( d(\cdot) \) denotes the depth function, which computes the depth of the static point \(\mathbf{T}_{\text{static}}'(i,t)\) after it has been transformed into the camera coordinate system at timestep \(t\), $\mathbf{D}_{\text{static}}(i,t)$ means the depth value for the $i$-th static point on time step $t$.
The total loss is defined as,
\begin{equation}
\mathcal{L}_{\text{static}} 
= \lambda_{\text{ba}} \mathcal{L}_{\text{ba}} 
+ \lambda_{\text{dc}} \mathcal{L}_{\text{dc}} 
+ \lambda_{\text{asap}} \mathcal{L}_{\text{asap}},
\end{equation}
where the weights are set as follows for all datasets: $\lambda_{\text{ba}} = 1$, $\lambda_{\text{dc}} = 1$, and $\lambda_{\text{asap}} = 5$.

In dynamic object tracking (Stage 3), the as-rigid-as-possible loss $\mathcal{L}_{\text{arap}}$~\cite{wang2024shape,yao2025uni4d} is used to constrain geometric deformation and prevent extreme shape changes. Specifically, for each dynamic control point \(k\) in $\mathbf{T}_{\text{dynamic}}$, we apply KNN to find its nearest neighbors among other tracks and enforce that the relative distances between these neighboring points remain consistent over time. The loss is formulated as:
\begin{equation}
\mathcal{L}_{\text{arap}} = \sum_{t=1}^{T} \sum_{k} \sum_{j \in \mathcal{N}(k)} \left\| \left( \mathbf{T}_{\text{dynamic}}(k,t) - \mathbf{T}_{\text{dynamic}}(j,t) \right) - \left( \mathbf{T}_{\text{dynamic}}(k,t-1) - \mathbf{T}_{\text{dynamic}}(j,t-1) \right) \right\|_2^2,
\end{equation}
where \(\mathcal{N}(k)\) denotes the set of nearest neighbors of control point \(k\), and \(\mathbf{T}_{\text{dynamic}}(k,t)\) is the position of point \(k\) at timestep \(t\).
In addition to the geometric constraint, we also introduce a temporal constraint—temporal smoothness loss $\mathcal{L}_{\text{ts}}$~\cite{yao2025uni4d}—to penalize abrupt changes across frames and ensure temporal coherence. It is defined as:
\begin{equation}
\mathcal{L}_{\text{ts}} = \sum_{t=1}^{T} \sum_{k} \left\| \mathbf{T}_{\text{dynamic}}(k,t) - \mathbf{T}_{\text{dynamic}}(k,t-1) \right\|_2^2,
\end{equation}
which enforces first-order smoothness in the trajectories of dynamic control points.
The total loss is defined as:
\begin{equation}
\mathcal{L}_{\text{dyn}} 
= \lambda_{\text{ba}} \mathcal{L}_{\text{ba}} 
+ \lambda_{\text{dc}} \mathcal{L}_{\text{dc}} 
+ \lambda_{\text{arap}} \mathcal{L}_{\text{arap}} 
+ \lambda_{\text{ts}} \mathcal{L}_{\text{ts}},
\end{equation}
where the weights are set as follows for all datasets:
$\lambda_{\text{ba}} = 1$, $\lambda_{\text{dc}} = 1$, $\lambda_{\text{arap}} = 100$, $\lambda_{\text{ts}} = 10$.

\subsection{Speeding-up the optimization} 

Directly using all tracks in camera pose optimization can be computationally prohibitive due to their large quantity. To mitigate this issue, we propose a simple strategy that reduces optimization overhead while preserving the final trajectory density. Specifically, we apply downsampling to the static tracking points while keeping the dynamic points intact. The rationale is twofold:

(1) Optimizing static points involves jointly estimating both camera poses and static trajectories, which is computationally more expensive. In contrast, optimizing dynamic points only requires recovering dynamic trajectories, incurring a much lower cost.

(2) Dynamic motion is of primary importance for understanding the scene. Downsampling dynamic points may cause the loss of fine-grained motion details, which we aim to preserve.

Specifically, we first downsample the static tracking points on the image plane by a factor of \( \frac{1}{\varpi^2} \). Let \( t \) denote the timestep, and let its coordinate in that frame be \( \mathbf{P}_{\text{static}}(i, t) \). We perform downsampling by dividing both the \( x \) and \( y \) components of \( \mathbf{P}_{\text{static}}(i, t) \) by \( \varpi \) and rounding them to the nearest integers:
\begin{equation}
\mathbf{P}^{\text{down}}_{\text{static}}(i, t) = \texttt{round}\left( \frac{\mathbf{P}^x_{\text{static}}(i, t)}{\varpi}, \frac{\mathbf{P}^y_{\text{static}}(i, t)}{\varpi} \right),
\end{equation}
where \( \texttt{round}(\cdot) \) denotes rounding to the nearest integer.
We then concatenate the initial frame index \( t \) with the downsampled coordinate to form a spatiotemporal key \( (t, \mathbf{P}^{\text{down}}_{\text{static}}(i, t)) \). A \(\texttt{unique}\) operation is applied to this set of keys to eliminate duplicates and retain only unique spatiotemporal locations. As a result, we obtain a reduced yet representative set of tracking points \( \mathbf{P}^{\text{down}}_{\text{static}} \) for subsequent optimization.
Next, we optimize the downsampled tracking points with the procedures described in the main paper. Upon completion of this optimization, we obtain the world-centric tracking points \( \mathbf{T}^{\text{down}}_{\text{static}} \), and then perform an upsampling process to densify the tracking points and reconstruct the full-resolution trajectories.
Specifically, for each tracking point \( i \), we consider its position on the image plane in the initial frame, denoted as \( \mathbf{P}_{\text{static}}(i, t) \). Using the associated depth value \( \mathbf{D}_{\text{static}}(i, t) \) and the estimated camera intrinsics, we back-project this point into the camera-centric 3D space to obtain \( \mathbf{P}^{3d}_{\text{static}}(i,t) \). Similarly, the downsampled point \( \mathbf{P}^{\text{down}}_{\text{static}} \) can be back-projected into 3D space to yield \( \mathbf{P}^{\text{down},3d}_{\text{static}} \).

\begin{figure}[!tbp]
  \centering
  \includegraphics[width=0.48\textwidth]{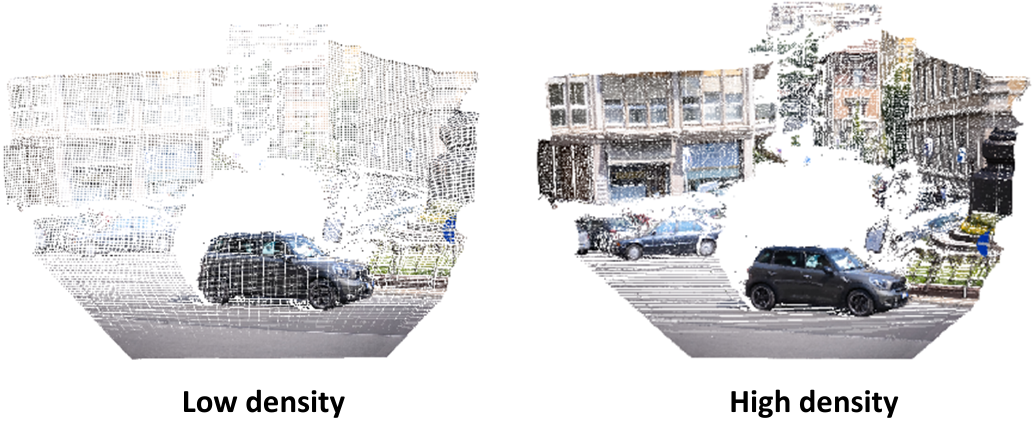}
  \caption{\textbf{Effectiveness of the speeding-up strategy.}}
  \label{low_high_density}
\end{figure}

To facilitate the upsampling, we search for the \( r \) nearest neighbors of \( \mathbf{P}^{3d}_{\text{static}}(i,t) \) among the downsampled 3D points \( \mathbf{P}^{\text{down},3d}_{\text{static}} \). Importantly, we restrict the search to those downsampled points that are visible in the current frame. Once the nearest neighbors are identified, we perform inverse-distance weighted interpolation to reconstruct the high-resolution trajectories. Concretely, we first compute the indices and distances of the neighbors using the \(\texttt{knn}\) module:
\begin{equation}
  \texttt{idx}, \texttt{dists} = \texttt{knn}(\mathbf{P}^{3d}_{\text{static}}(i,t), \mathbf{P}^{\text{down},3d}_{\text{static}}, K=r+1), 
\end{equation}
where \( K = r + 1 \) to include the query point itself.
We then compute the interpolation weights as follows:
\begin{align}
w_x &= \frac{1}{\texttt{dists}_x + \epsilon}, \\
\hat{w}_x &= \frac{w_x}{\sum_y w_y},
\end{align}
where \( \epsilon \) is a small constant added for numerical stability.
Finally, we recover the full-resolution world-centric trajectories by computing a weighted aggregation over the downsampled points:
\begin{equation}
 \mathbf{T}^{\text{full}}_{\text{static}} = \mathcal{F} (\hat{w}_x, \mathbf{T}^{\text{down}}_{\text{static}}),
\end{equation}
where \( \mathcal{F} \) denotes the weighted aggregation function.
The visualization of the proposed speeding-up strategy is shown in Fig.~\ref{low_high_density}, and its quantitative effectiveness on the Sintel dataset is reported in Table~\ref{table:depth}.

\begin{table}[!ht]
  \begin{center}
    \footnotesize
    \caption{\textbf{Quantitative evaluation of the speeding-up strategy on the Sintel dataset.} The proposed method achieves similar accuracy with reduced optimization time. }
    \label{table:depth}
    \scalebox{1}{\begin{tabular}{c|ccc|cc|c}
      \toprule
       \multirow{2}{*}{speeding-up strategy}&\multicolumn{6}{c}{Sintel} \\
       & ATE $\downarrow$&RTE$\downarrow$   & RRE$\downarrow$& Abs Rel $\downarrow$& $\delta<1.25 \uparrow$ & T (min)\\
       \midrule
       
        \ding{55}&0.089&\textbf{0.034}&0.414&\textbf{0.207}&\textbf{73.8}&60\\
        \ding{51}&\textbf{0.088}&0.035&\textbf{0.410}&0.218&73.3& \textbf{8}\\
        
      \bottomrule
    \end{tabular}}
     \vspace{-1.2em}
  \end{center}
\end{table}

\subsection{Consistent Video Depth Generation.}
Our method is capable of producing temporally consistent video depth, as it performs dense 3D tracking for nearly all pixels. This enables the generation of consistent depth sequences through a straightforward interpolation-based propagation step. 

Assume we have obtained $M$ dense 3D tracking points across a video sequence. For each frame $t \in \{1, 2, \dots, T\}$, let $\mathcal{M}_t$ denote the set of visible 3D tracking points and $D_t$ their corresponding camera-centric depths. We also denote the raw monocular depth prediction for frame $t$ as $D_t^{\text{raw}}$.

To align the raw monocular depth with our optimized tracking-based depth, we compute a per-point local scale ratio rather than a global frame-wise scale. For each tracking point $\mathbf{M}_t^i \in \mathcal{M}_t$ with image-plane projection $\mathbf{p}_t^i$, the scale ratio is defined as:
\begin{equation}
r_t^i = \frac{D_t(\mathbf{p}_t^i)}{D_t^{\text{raw}}(\mathbf{p}_t^i)}.
\end{equation}

For an arbitrary pixel $\mathbf{p}_t$ in frame $t$, its final aligned depth $\hat{D}_t(\mathbf{p}_t)$ is obtained by propagating the local scale information from nearby 3D tracking points. Let $\mathcal{N}_t(\mathbf{p}_t) \subset \mathcal{M}_t$ denote its $k$ nearest neighbors in the image plane. For each neighbor $\mathbf{M}_t^j \in \mathcal{N}_t(\mathbf{p}_t)$, we compute the 3D distance based on their raw-depth-lifted coordinates:
\begin{equation}
\mathbf{P}_t = D_t^{\text{raw}}(\mathbf{p}_t) \cdot K^{-1} \tilde{\mathbf{p}}_t, \quad 
\mathbf{Q}_t^j = D_t^{\text{raw}}(\mathbf{p}_t^j) \cdot K^{-1} \tilde{\mathbf{p}}_t^j,
\end{equation}
\begin{equation}
d_t^j = \| \mathbf{P}_t - \mathbf{Q}_t^j \|_2,
\end{equation}
where $K$ is the camera intrinsic matrix and $\tilde{\mathbf{p}}$ denotes the homogeneous coordinate of pixel $\mathbf{p}$.

We assign each neighbor an inverse-distance weight:
\begin{equation}
w_t^j = \frac{1}{d_t^j + \epsilon}, \quad 
\tilde{w}_t^j = \frac{w_t^j}{\sum_{j=1}^k w_t^j},
\end{equation}
where $\epsilon$ is a small constant for numerical stability. The interpolated local scale ratio at pixel $\mathbf{p}_t$ is then computed as:
\begin{equation}
r_{\mathbf{p}_t} = \sum_{j=1}^k \tilde{w}_t^j \cdot r_t^j,
\end{equation}
and the final aligned depth is given by:
\begin{equation}
\hat{D}_t(\mathbf{p}_t) = r_{\mathbf{p}_t} \cdot D_t^{\text{raw}}(\mathbf{p}_t).
\end{equation}

Through this weighted propagation, accurate scale information from the dense 3D tracks is effectively diffused across the entire image, producing a temporally consistent and spatially coherent video depth sequence. As shown in Tab.~\ref{table:videodepth}, 

\begin{table}[!ht]
  \begin{center}
    \footnotesize
    \setlength\tabcolsep{1.5pt}
    \caption{\textbf{Video depth estimation results.} We evaluate our model on three datasets: Sintel, Bonn and TUM D. \textbf{Best} results are highlighted.}
    \label{table:videodepth}
    \scalebox{1}{\begin{tabular}{ll|cc|cc|cc}
      \toprule
       \multirow{2}{*}{Category} &\multirow{2}{*}{Method}&\multicolumn{2}{c|}{Sintel} &\multicolumn{2}{c|}{Bonn} &  \multicolumn{2}{c}{TUM D}    \\
       & & Abs Rel $\downarrow$& $\delta<1.25 \uparrow$   & Abs Rel $\downarrow$& $\delta<1.25 \uparrow$& Abs Rel $\downarrow$& $\delta<1.25 \uparrow$\\
       \midrule
       &Depth Anything V2~\cite{yang2024depthanythingv2}&0.348&59.2&0.106 &92.1&0.211&78.0\\
       Single-frame&Depth Pro~\cite{bochkovskii2024depthpro}&0.418&55.9&0.068&\textbf{97.4}&0.126&89.3\\
       depth&ZoeDepth~\cite{bhat2023zoedepth}&0.467 &47.3&0.087 &94.8&0.176&74.5\\
       &Unidepth~\cite{piccinelli2024unidepth}&0.473& 63.0 &0.057& 97.4& 0.113& 91.9\\
       \midrule
       \multirow{2}{*}{Video depth} &ChronoDepth~\cite{shao2024chronodepth}& 0.687 &48.6 &0.100& 91.1&/&/\\
       &DepthCrafter~\cite{hu2024depthcrafter}&0.292 &69.7&0.075& 97.1&/&/\\
       \midrule
       &DUSt3R~\cite{wang2024dust3r}&0.422&54.2&0.144&84.5&0.239&71.1\\
        Joint video  &MonST3R~\cite{zhang2024monst3r}&0.335&58.6&0.063& 96.4&0.301&55.8\\
        depth \& pose &Align3R (Depth Pro)~\cite{lu2024align3r}&0.263&64.1&0.058&97.1&0.111&88.9\\
        &Ours (DELTA~\cite{ngo2024delta})&\textbf{0.222}&\textbf{72.6}&0.058&97.3&\textbf{0.086}&\textbf{92.3}\\
        &Ours (CoTrackerV3~\cite{karaev2024cotracker3})&0.232&71.4&\textbf{0.054}&97.3&0.090&91.7\\
        
      \bottomrule
    \end{tabular}}
     \vspace{-1.2em}
  \end{center}
\end{table}

\subsection{Runtime comparison with baseline}

To further evaluate the efficiency and effectiveness of our optimization strategy, we compare our method with a strong baseline that combines camera poses and consistent depths from Uni-4D~\cite{yao2025uni4d} with dense camera-centric 3D tracking from DELTA. Since both methods share the same 3D tracking backbone, the comparison focuses on optimization runtime and accuracy.

Uni4D estimates camera poses in a streaming manner, where the pose between consecutive frames is computed step by step. This results in a total runtime that scales linearly with the video length. In contrast, our approach performs pose estimation in a clip-to-global parallel manner, significantly improving computational efficiency.

In terms of 3D tracking accuracy, our method achieves higher precision by effectively distinguishing static and dynamic regions using dynamic masks, and by jointly optimizing both camera poses and 3D trajectories. This joint optimization leads to improved geometric consistency and reconstruction quality.

We report detailed quantitative results on camera pose estimation (Sintel, frames 30–50) and world-coordinate 3D tracking (ADT, first 64 frames) in Tab.~\ref{table:runtime}. All experiments are conducted on the same hardware configuration for a fair comparison. The results demonstrate that our method achieves both higher accuracy and lower runtime compared to the Uni4D baseline. The improvement mainly stems from our clip-to-global parallel optimization strategy and the integration of dynamic mask filtering, which together enhance efficiency and reconstruction quality.

\begin{table}[!ht]
\centering
\caption{Runtime and accuracy comparison on Sintel (30–50 frames) and ADT (first 64 frames).}
\label{table:runtime}
\scalebox{0.9}{\begin{tabular}{lcccccc}
\toprule
\multirow{2}{*}{Setting} & \multicolumn{4}{c}{Sintel} & \multicolumn{2}{c}{ADT } \\
\cmidrule(lr){2-5} \cmidrule(lr){6-7}
 & ATE ↓ & RTE ↓ & RPE ↓ & Avg. Time (min) ↓ & APD$_{3D}$ ↑ & Avg. Time (min) ↓ \\
\midrule
Uni4D + DELTA & 0.118 & 0.048 & 0.610 & 19 & 68.95 & 28 \\
\textbf{Ours (DELTA)} & \textbf{0.087} & \textbf{0.036} & \textbf{0.406} & \textbf{15} & \textbf{75.18} & \textbf{20} \\
\bottomrule
\end{tabular}}
\end{table}

\subsection{More ablation study} 

\paragraph{Ablation on the filtering mechanism.}
\label{filtering}
In this section, we introduce the details of the filtering process. We first compute the complement set by discarding pixels that lie in any previously visible 2D track trajectory. However, this operation may introduce isolated pixels, which are typically not meaningful for downstream scene understanding. To mitigate this, we construct connected components from the newly selected 2D tracking points and apply a size threshold~$\tau$ to remove small components. Only connected regions with more than $\tau$ pixels are retained. This ensures that the preserved 2D tracks are geometrically meaningful and more likely to correspond to coherent object parts.
We found that this filtering procedure consistently improves accuracy, as most filtered points are either outliers or redundantly close to already tracked regions. Moreover, the threshold~$\tau$ is robust across different scenes: we use a fixed value of $\tau = 50$ in all experiments without additional tuning.

\begin{table}[!ht]
\centering
\footnotesize
\caption{Ablation study on the filtering mechanism on Sintel and Bonn datasets. Filtering improves both camera pose estimation and tracking stability.}
\vspace{0.5em}
\begin{tabular}{lcccccc}
\toprule
\multirow{2}{*}{Setting} & \multicolumn{3}{c}{Sintel} & \multicolumn{3}{c}{Bonn} \\
\cmidrule(lr){2-4} \cmidrule(lr){5-7}
 & ATE $\downarrow$ & RTE $\downarrow$ & RPE $\downarrow$ & ATE $\downarrow$ & RTE $\downarrow$ & RPE $\downarrow$ \\
\midrule
w/o Filtering & 0.105 & 0.038 & 0.442 & 0.018 & 0.007 & 0.601 \\
w/ Filtering  & \textbf{0.088} & \textbf{0.035} & \textbf{0.410} & \textbf{0.016} & \textbf{0.005} & \textbf{0.564} \\
\bottomrule
\end{tabular}
\end{table}

\subsection{More visualization} 
\subsubsection{Comparison on camera pose estimation}
\begin{figure}[!t]
    \begin{center}
        \includegraphics[width=1\textwidth]{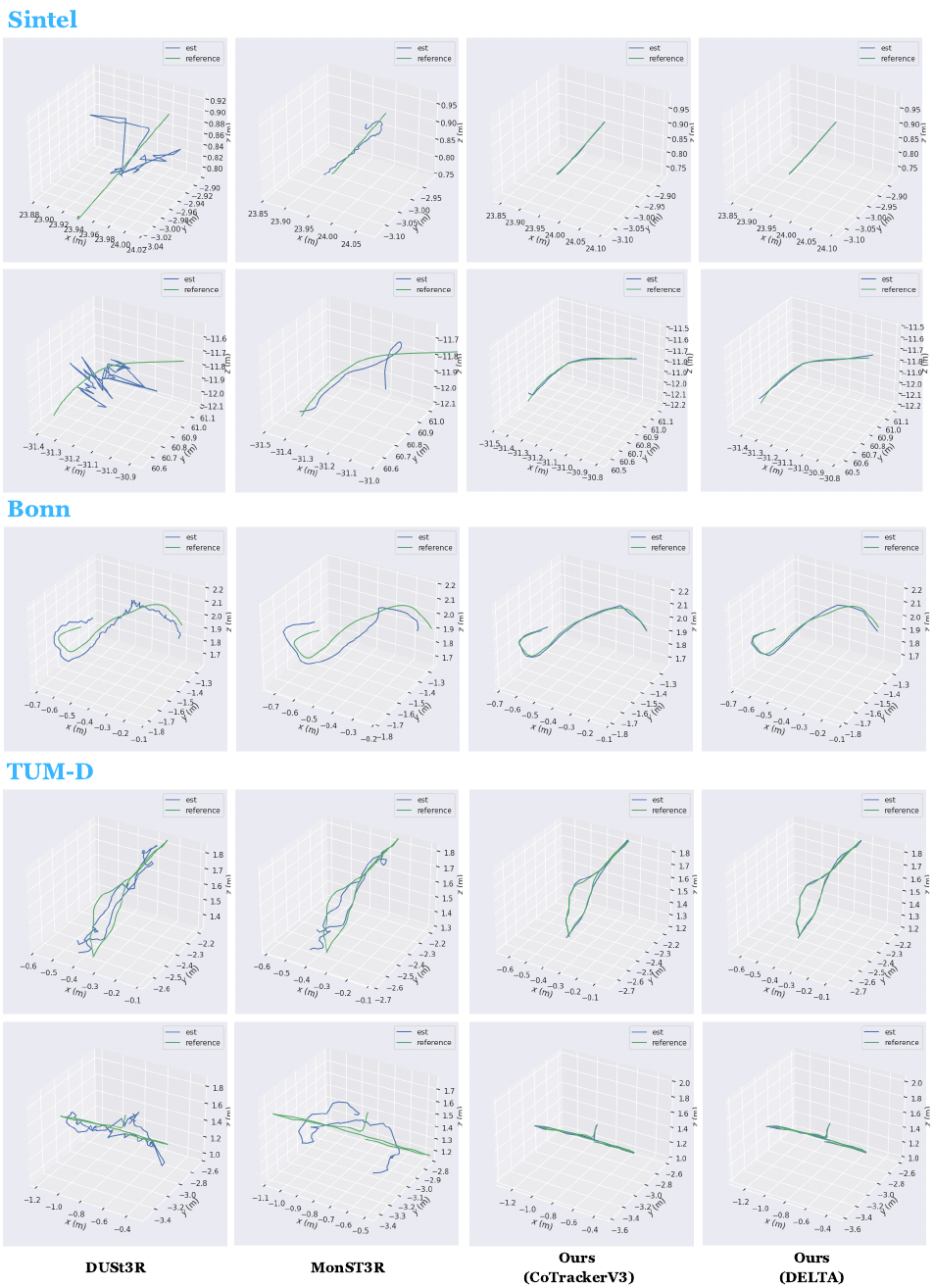}
        \caption{\textbf{Camera pose estimation comparison} on the Sintel~\cite{butler2012naturalistic} 
        Bonn~\cite{palazzolo2019refusion} and TUM-D~\cite{sturm2012benchmark} datasets.}
        \label{figure:pose_curve}
    \end{center}
\end{figure}

Fig.~\ref{figure:pose_curve} illustrates qualitative comparisons of camera pose estimation on the Sintel~\cite{butler2012naturalistic}, Bonn~\cite{palazzolo2019refusion}, and TUM-D~\cite{sturm2012benchmark} datasets. We evaluate our method against two joint depth and pose estimation baselines: DUSt3R~\cite{wang2024dust3r} and MonST3R~\cite{zhang2024monst3r}. As shown, our method produces camera trajectories that are more stable and better aligned with the ground truth, reflecting enhanced robustness and accuracy.

\subsubsection{World-centric dense tracking results }
\begin{figure}[!t]
    \begin{center}
        \includegraphics[width=1\textwidth]{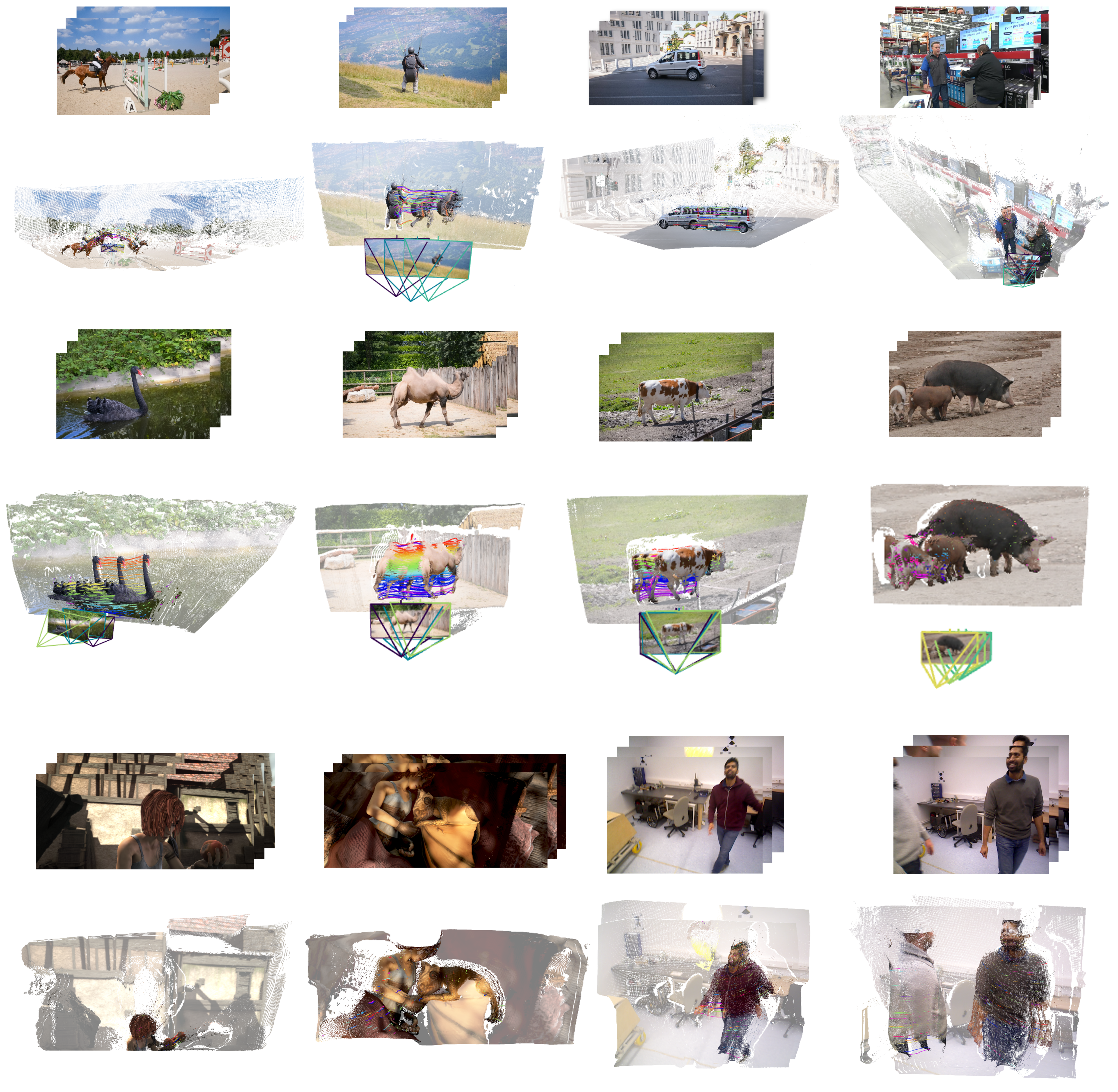}
        \caption{\textbf{More Qualitative results.} Our method can output 3D tracks in a world-centric coordinate system. 
        }
        \label{vis_sup}
    \end{center}
\end{figure}

We present additional visualizations of the world-centric tracking results in Fig.~\ref{vis_sup}. To enhance clarity, we only visualize the point clouds on temporally spaced keyframes. However, the displayed trajectories are computed by connecting 3D tracks across all frames, capturing the complete motion over time. For more vivid and dynamic results, please refer to the accompanying video in the supplementary material.

\subsection{Limitation and future work}
Our method currently relies on several auxiliary models to obtain 2D tracks, monocular depths, and dynamic masks. This dependence introduces additional computational overhead and imposes stringent quality requirements on these components. In the future, feed-forward solutions may offer a more suitable and efficient direction. For instance, St4RTrack~\cite{feng2025st4rtrack} adopts a feed-forward design, but its pair-wise matching strategy inherently suffers from drift accumulation, which requires global optimization for correction. Inspired by VGGT~\cite{wang2025vggt}, a promising direction may involve jointly processing all frames to directly predict the state of each frame across time. This could potentially enable a more consistent and globally coherent trajectory estimation.

\subsection{Assets availability}
\label{asset}
The datasets used in this study and their respective licenses are listed below:

\begin{itemize}
    \item \textbf{Sintel}~\cite{butler2012naturalistic}: Available at \url{http://sintel.is.tue.mpg.de/}. This dataset is intended for optical flow evaluation. Please refer to the official website for specific license information.
    
    \item \textbf{Bonn RGB-D Dynamic Dataset}~\cite{palazzolo2019refusion}: Available at \url{https://www.ipb.uni-bonn.de/data/rgbd-dynamic-dataset/}, licensed under the Creative Commons Attribution-NonCommercial-ShareAlike 3.0 Unported License.
    
    \item \textbf{TUM RGB-D Dataset}~\cite{sturm2012benchmark}: Available at \url{https://cvg.cit.tum.de/data/datasets/rgbd-dataset}, licensed under the Creative Commons Attribution 4.0 International License (CC BY 4.0).
    
    \item \textbf{DAVIS}~\cite{perazzi2016benchmark}: Available at \url{https://davischallenge.org/}. According to the DAVIS 2017 Challenge, the dataset is licensed under the Creative Commons Attribution 4.0 License.
    
    \item \textbf{Aria Digital Twin (ADT)}~\cite{pan2023aria}: Available at \url{https://www.projectaria.com/datasets/adt/}. Provided by Meta Reality Labs Research; please consult the official website for license details.
    
    \item \textbf{Panoptic Studio Dataset}~\cite{joo2015panoptic}: Available at \url{https://www.cs.cmu.edu/~hanbyulj/panoptic-studio/}. Provided by Carnegie Mellon University; please refer to the project website for license information.
    
    \item \textbf{CVO Dataset}~\cite{wu2023accflow}: Available at \url{https://github.com/mulns/AccFlow/blob/main/data/README.md}, licensed under the MIT License.
\end{itemize}

\end{document}